\documentclass[review]{elsarticle}
\usepackage{lineno,hyperref}
\modulolinenumbers[5]

\makeatletter
\def\ps@pprintTitle{%
 \let\@oddhead\@empty
 \let\@evenhead\@empty
 \def\@oddfoot{}%
 \let\@evenfoot\@oddfoot}
\makeatother

\usepackage[utf8]{inputenc}  
\usepackage{listings}
\usepackage{color}
\usepackage{amssymb}  
\usepackage{amsmath}  
\usepackage{mathrsfs} 
\usepackage{bm}  
\usepackage{subcaption}  
\usepackage{setspace}
\usepackage[nolist,nohyperlinks]{acronym}

\usepackage[margin=1.5cm]{geometry}

\biboptions{sort&compress} 

\usepackage{array}
\newcolumntype{L}[1]{>{\raggedright\let\newline\\\arraybackslash\hspace{0pt}}m{#1}}
\newcolumntype{C}[1]{>{\centering\let\newline\\\arraybackslash\hspace{0pt}}m{#1}}
\newcolumntype{R}[1]{>{\raggedleft\let\newline\\\arraybackslash\hspace{0pt}}m{#1}}
\newcolumntype{X}[1]{>{\raggedright\let\newline\\\arraybackslash\hspace{0pt}}p{#1}}
\newcolumntype{Y}[1]{>{\centering\let\newline\\\arraybackslash\hspace{0pt}}p{#1}}
\newcolumntype{Z}[1]{>{\raggedleft\let\newline\\\arraybackslash\hspace{0pt}}p{#1}}

\usepackage{siunitx}  
\definecolor{cadmiumgreen}{rgb}{0.0, 0.42, 0.24}

\lstdefinestyle{floatcode}{float=tp,floatplacement=tbp
}

\newcommand\changed[1]{#1}

\newsavebox\CBox

\newcommand\descriptor{\mathcal{D}}
\newcommand\comparator{\mathcal{C}}
\newcommand\ranker{R}
\newcommand\rank{\tau}
\newcommand\rankof[1]{\rank_{#1}}
\newcommand\rankposition[2]{\rho_{#1}(#2)}

\newcommand\ranksof[1]{\mathcal{T}_{#1}}
\newcommand\scoresymbol{\varsigma}

\newcommand\scorein[3]{\varsigma_{#1}(#2,#3)}

\newcommand\FG{\mathcal{G}}
\newcommand\FGset{\mathbb{G}}
\newcommand\FGProjector{\mathcal{E}}
\newcommand\FV{\mathcal{V}}
\newcommand\FVset{\mathbb{V}}
\newcommand\real{\rm I\!R}

\newacro{ACC}{Auto color correlogram}
\newacro{BIC}{Border/Interior Pixel Classification}
\newacro{BoG}{Bag of Graphs}
\newacro{CEDD}{Color and Edge Directivity Descriptor Spatial Pyramid}
\newacro{CCOM}{Color Co-Occurrence Matrix}
\newacro{FCTH}{Fuzzy Color and Texture Histogram Spatial Pyramid}
\newacro{GCH}{Global Color Histogram}
\newacro{GoI}{Graph of Interest}
\newacro{JAC}{Joint Autocorrelogram}
\newacro{JCD}{Joint Composite Descriptor}

\journal{Pattern Recognition Letters}
\begin{document}

\sloppy  

\begin{frontmatter}

\title{Multimodal Prediction based on Graph Representations}

\author[address1]{Icaro Cavalcante Dourado\corref{correspondingauthor}}
\ead{icaro.dourado@ic.unicamp.br}

\author[address2]{Salvatore Tabbone}
\ead{tabbone@loria.fr}

\author[address3]{Ricardo da Silva Torres}
\ead{ricardo.torres@ntnu.no}

\cortext[correspondingauthor]{Corresponding author. Av. Albert Einstein, 1251, Campinas, SP, Brazil.}

\address[address1]{Institute of Computing, University of Campinas (UNICAMP), Campinas, Brazil}

\address[address2]{Universit\'e de Lorraine-LORIA UMR 7503, Vandoeuvre-l\`es-Nancy, France}

\address[address3]{Department of ICT and Natural Sciences, Norwegian University of Science and Technology (NTNU), {\AA}lesund, Norway}

\begin{abstract}
\changed{This paper proposes a learning model, based on rank-fusion graphs, for general applicability in multimodal prediction tasks, such as multimodal regression and image classification.
Rank-fusion graphs encode information from multiple descriptors and retrieval models, thus being able to capture underlying relationships between modalities, samples, and the collection itself.
The solution is based on the encoding of multiple ranks for a query (or test sample), defined according to different criteria, into a graph. Later, we project the generated graph into an induced vector space, creating fusion vectors, targeting broader generality and efficiency. A fusion vector estimator is then built to infer whether a multimodal input object refers to a class or not.
Our method is capable of promoting a fusion model better than early-fusion and late-fusion alternatives.
Performed experiments in the context of multiple multimodal and visual datasets, as well as several descriptors and retrieval models, demonstrate that our learning model is highly effective for different prediction scenarios involving visual, textual, and multimodal features, yielding better effectiveness than state-of-the-art methods.}
\end{abstract}

\begin{keyword}
graph-based rank fusion \sep graph embedding \sep multimodal fusion \sep prediction tasks \sep rank aggregation \sep representation model
\end{keyword}

\end{frontmatter}

\section{Introduction}

Nowadays, data analysis involving multimedia and heterogeneous content is a hot topic that attracts a lot of attention from not only public and private sectors, but also academia.
The proliferation of digital content and social media is expanding substantially the volume and diversity of available digital content. Most of these data are unlabeled, heterogeneous, unstructured, and derived from multiple modalities.

Despite such challenges, such content is of great relevance to support the development of prediction and retrieval models. In particular,
multimodal data analysis is required in several scenarios, such as content-based information retrieval, and multimedia event detection of natural disasters, such as flooding. 


Massive digital content has demanded the development of textual, visual, and multimedia descriptors for content-based data analysis.
Despite the continuous advance on feature extractors and machine learning techniques, a single descriptor or a single modality is often insufficient to achieve effective prediction results in real case scenarios.
Descriptors have specific pros and cons because each one often focuses on a specific point of view of a single modality. For example, dedicated descriptors may be created to characterize scenes, textual descriptions, movement, symbols, signals, etc. For this reason, descriptors and retrieval models often provide complementary views, when adopted in combination.

Many works have been proposed to combine heterogeneous data sources (remotely sensed information and social media) to promote multimodal analysis. In~\cite{zhou:2017:CBIR}, authors point out the benefit of exploring and fusing multimodal features with different models. Moreover, combining different kinds of features (local vs holistic) improves substantially the retrieval effectiveness~\cite{zhang:2015:queryRankFusion}, and most search fusion approaches are based on rank fusion~\cite{dao:2018,Dourado:2019:FG,zhang:2015:queryRankFusion}.

Scenarios involving heterogeneous data impose a challenge of selecting features or models to combine, which is performed by either supervised or unsupervised approaches. Unsupervised approaches are necessary in the absence of labeled data, which is prominent nowadays, or in scenarios involving lower computation capabilities or a large amount of data.

Most previous initiatives for multimodal prediction are solely based on either CNN-based descriptors in isolation, feature concatenation~\cite{dirsm17:BBischkeEtAl,dirsm17:MSDaoEtAl,dirsm17:XFuEtAl,dirsm17:MHanifEtAl,dirsm17:LLopezFuentesEtAl,dirsm17:NTkachenkoEtAl}, or graph-based feature fusion~\cite{werneck:2018:BocgBokg,dirsm17:KAvgerinakisEtAl}.
These approaches still ignore the correlation between modalities, as well as object correlation, and they are not consistently better than ranking models that do not rely on fusion.

\changed{This paper proposes a learning model, based on rank-fusion graphs, for general applicability in multimodal prediction tasks, such as classification and regression.
We explore and extend the concept of a rank-fusion graph -- originally proposed as part of a rank aggregation approach for retrieval tasks~\cite{Dourado:2019:FG} -- for supervised learning and additional applications.}
Our fusion method relies on the representation of multiple ranks, defined according to different criteria, into a graph. Graphs provide an efficient representation of arbitrary structures and inter-relationships among different elements of a model.
We embed the generated graph into a feature space, creating fusion vectors. Next, an estimator is trained to predict if an input multimodal object refers to a target label (or event) or not, following their fusion vectors.

In~\cite{Dourado:2019:FV}, a methodology to apply rank-fusion graphs for efficient retrieval was presented, in the context of retrieval tasks.
\changed{Different from that work, here we propose a new learning paradigm over fusion vectors for multimodal prediction.
We propose specific components for the training and prediction phases, as well as additional applications for the fusion vectors.}

Our method performs unsupervised learning for the object representation, exploring and capturing relationships from the collection into the representation model. It works on top of any descriptors for multimodal data, such as visual or textual. By promoting a representation model solely based on base descriptors and unsupervised data analysis over the collection, we conjecture that our approach leads to a competitive multimodal representation model that explores and encodes information from multiple descriptors and underlying sample relationships automatically, while not requiring labeled data.

Experimental results over multiple multimodal and visual datasets demonstrate that the proposal is robust for different detection scenarios involving textual, visual, and multimodal features, yielding better detection results than state-of-the-art methods from both early-fusion and late-fusion approaches.

\changed{This paper extends the work presented in~\cite{Dourado:2019:fusionGraphPrediction} where the notion of a representation model from rank-fusion graphs was introduced and evaluated for multimedia flood detection.
Here we extend this approach, proposing and discussing alternative approaches for the representation model.
We also show additional applications for it, promoting general applicability for prediction tasks. The method is evaluated over multiple multimodal and image scenarios.
We also evaluate the method against additional baselines, either early-fusion and late-fusion approaches.
Finally, we evaluate the influence of hyper-parameters and provide guidelines for their selection.}

\section{Related Work}

Representation learning models have been developed and advanced for the data modalities individually, such as image, text, video, and audio.
However, their combined exploration for multimodal tasks remains an open issue.
Multimodality imposes even more challenges, depending on the scenario, such as translation between modalities, exploration of complementarity and redundancy, co-learning, and semantic alignment~\cite{baltruvsaitis:2018:multimodal}.

Some works have focused on multimodal events, which require modeling spatio-temporal characteristics of data~\cite{tzelepis2016event}. Faria et al.~\cite{faria2016fusion} proposed a time series descriptor that generates recurrence plots for the series, coupled with a bio-inspired optimization of how to combine classifiers. In this paper, we focus on multimodal tasks that do not depend on temporal modeling as well as unsupervised models.

In order to achieve fusion capabilities, \textit{early-fusion} approaches emphasize the generation of composite descriptions for samples, thus working at the feature level.
Conversely, \textit{late-fusion} approaches perform a combination of techniques focused on a target problem, fusing at score or decision levels.
On a smaller scale, some papers propose hybrid solutions based on both approaches~\cite{zadeh2017tensor,lan2014multimedia}.
As we propose a representation model, our solution can be seen as an early-fusion approach. Nevertheless, it is based on retrieval models, without the need to work directly on a feature level. In this sense, we categorize the method as a hybrid.

While early-fusion approaches are theoretically able to capture correlations between modalities, often a certain modality produces unsatisfactory performance and leads to biased or over-fitted models~\cite{zadeh2017tensor}.
Most early-fusion methods work in a two-step procedure; first extracting features from different modalities, and then fusing them by strategies such as concatenation~\cite{kiela2014learning}, singular values decomposition~\cite{bruni2014multimodal}, or autoencoders~\cite{silberer2014learning}. A few others focus on multimodal features jointly~\cite{kottur2016visual,hill2014learning}, although generally restricted to a pre-defined textual attribute set.
Concatenation is a straightforward yet widely used early-fusion approach, which merges vectors obtained by different descriptors. As a drawback, concatenation does not explore inherent correlations between modalities.

Supervised early-fusion optimizes a weighted feature combination, either during or after feature extraction.
A common strategy is to build a neural architecture with multiple separate input layers, then including a final supervised layer, such as a regressor~\cite{nogueira2018exploiting}.
Another approach is the design of a composite loss function, suited for the particular desired task~\cite{park2016image}. Composite loss functions work well in practice, but, as they need both multimodal composition and supervision, they are tied to the domain of interest.
Supervised early-fusion usually suffers from high memory and time consumption costs. Besides, they usually have difficulty in preserving feature-based similarities and semantic correlations~\cite{lan2014multimedia}.

Late-fusion approaches are useful when the raw data from the objects are not available. Besides, they are less prone to over-fit.
Mixture of experts (MoE) approaches focus on performing decision fusion, combining predictors to address a supervised learning problem~\cite{yuksel2012twenty}.
Majority voting of classifiers~\cite{nogueira2018exploiting}, rank aggregation functions~\cite{Dourado:2019:FG}, and matrix factorization~\cite{dong2018late} are examples of late-fusion methods.
Both fusions based on rank aggregation functions and matrix factorization are based on manifold learning, i.e., the exploration of dataset geometry.

Majority voting is a well-known approach to combine multiple estimators, being effective due to bias reduction. It is applied to scenarios involving an odd number of estimators so that each predicted output is taken as the one most frequently predicted by the base estimators.

Rank aggregation functions combine results from different retrieval models. They target a good permutation of retrieved objects obtained from different input ranks, in order to promote more effective retrieval results~\cite{Dourado:2019:FG}.

In general, unsupervised learning needs more investigation, especially for multimodal representation models. We explore how unsupervised rank aggregation capabilities can be applied to prediction tasks. Despite their original intent regarding better retrieval accuracy, we claim that unsupervised rank aggregation functions can provide effective dataset exploitation.

Considering previous works regarding fusion approaches for prediction tasks, a considerable amount of them are still based on classic visual descriptors~\cite{dirsm17:MSDaoEtAl,dirsm17:MHanifEtAl,dirsm17:NTkachenkoEtAl,werneck:2018:BocgBokg}.
Most of them resorted to pre-trained CNN-based models for visual feature extraction~\cite{dirsm17:KAhmadEtAl,dirsm17:SAhmadEtAl,dirsm17:KAvgerinakisEtAl,dirsm17:BBischkeEtAl,dirsm17:XFuEtAl,dirsm17:LLopezFuentesEtAl,dirsm17:KNogueiraEtAl}, from which just a few fine-tuned their models~\cite{dirsm17:LLopezFuentesEtAl,dirsm17:KNogueiraEtAl}.
When dealing with specific tasks, especially for competitions, some works use preprocessing steps, such as image cropping and filtering~\cite{dirsm17:ZZhao}, but this is beyond our general intent in this paper.  In order to explore the textual modality, most initiatives used BoW, using either TF or TF-IDF weighting, while others presented more complex formulations, such as word embeddings~\cite{dirsm17:BBischkeEtAl,dirsm17:NTkachenkoEtAl}, Long Short-Term Memory (LSTM) networks~\cite{dirsm17:LLopezFuentesEtAl}, or relation networks~\cite{dirsm17:KNogueiraEtAl}.
Regarding multimodal scenarios, most works rely on early-fusion approaches, such as a concatenation of visual and textual feature vectors~\cite{dirsm17:BBischkeEtAl,dirsm17:MSDaoEtAl,dirsm17:XFuEtAl,dirsm17:MHanifEtAl,dirsm17:LLopezFuentesEtAl,dirsm17:NTkachenkoEtAl} or graph-based attribute fusion~\cite{dirsm17:KAvgerinakisEtAl,werneck:2018:BocgBokg}, while only a few others adopted late-fusion approaches~\cite{dirsm17:KAhmadEtAl,dirsm17:SAhmadEtAl}.

\section{Preliminaries}
\label{secPrelim}

Let a \textit{sample} $s$ be any digital object, such as a document, an image, a video, or even a hybrid (multimodal) object.
$s$ is characterized by a \textit{descriptor} $\descriptor$, which relies on a particular point of view to describe $s$ as a vector, graph or any data structure $\epsilon$. Descriptions allow samples to be compared to each other, thus composing the basis of retrieval and learning models.

A \textit{comparator} $\comparator$, applied over a tuple $(\epsilon_i,\epsilon_j)$, produces a \textit{score} $\scoresymbol \in \mathbb{R^+}$ (e.g., the Euclidean distance or the cosine similarity). Either similarity or dissimilarity functions can be used to implement $\comparator$. 
A query sample, or just \textit{query} $q$, refers to a particular sample taken as an input object in the context of a search, whose purpose is to retrieve \textit{response items} from a \textit{response set} ($S$) according to relevance criteria. A response set ${S = \{ s_i\}}_{i=1}^n$ is a collection of $n$ \textit{samples}, where $n$ is the collection size.
In prediction tasks, a query is called a \textit{test sample}, and a response set is called a \textit{train set}. These terms can be used interchangeably, but a train set refers to labeled samples.

A \textit{ranker} is a tuple $\ranker = (\descriptor,\comparator)$, which is employed to compute a rank $\rank$ for $q$, denoted by $\rankof{q}$ to distinguish its corresponding query.
A rank is defined as a permutation of $S_L \subseteq S$, where $L \ll n$ in general, subject to $\rankof{q}$ providing the least dissimilar samples to $q$ from $S$, in order.
$L$ is a cut-off parameter.
\changed{The position of $x$ in $\rankof{q}$ is expressed by $\rankposition{\rankof{q}}{x}$, starting by value $1$.
$\scorein{\rankof{q}}{s_i}{s_j}$ is the score between $s_i$ and $s_j$ with respect to the same descriptor and comparator from the particular ranker that produced the rank $\rankof{q}$ for the query $q$.}

A ranker establishes a ranking system, but different descriptors and comparators can compose rankers. Besides, descriptors are commonly complementary, as well as comparators.
Given $m$ rankers, $\{R_j\}_{j=1}^m$, used for query retrieval over a collection $S$, for every query $q$, we can obtain $\ranksof{q}= \{\rankof{j}\}_{j=1}^m$, from which a \textit{rank aggregation} function $f$ produces a combined rank $\rankof{q,f}=f(\ranksof{q})$, presumably more effective than the individual ranks $\rankof{1}$, $\rankof{2}$, $\dots$, $\rankof{m}$.

\section{Representation and Prediction Based on Graph-Based Rank Fusion}
\label{secMethod}

\begin{figure}
\centering
\includegraphics[width=0.45\linewidth]{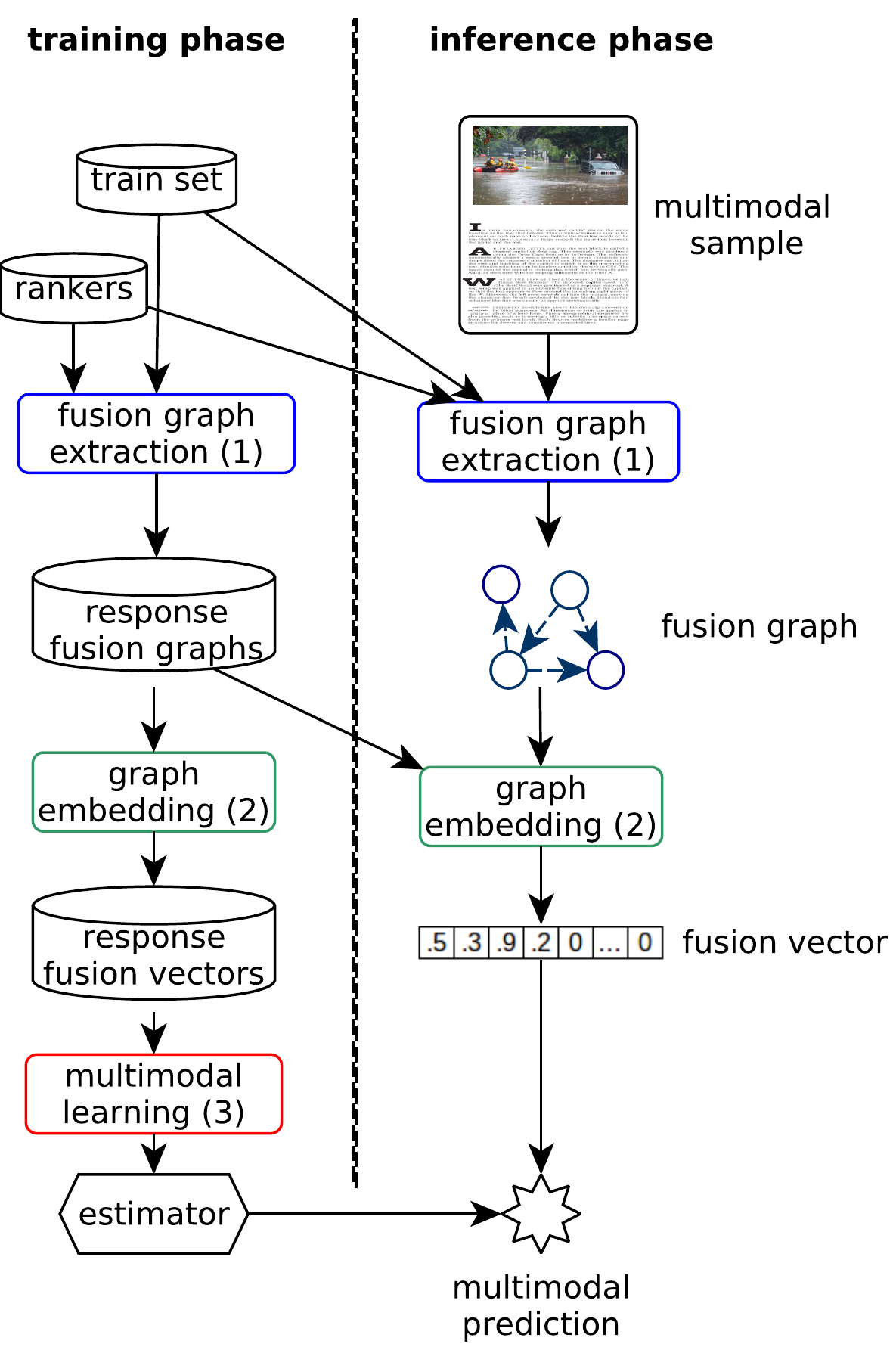}
\caption{Graph-based rank fusion for multimodal prediction.}\label{fig:fusionVectorEstimation}
\end{figure}

\changed{Fig.~\ref{fig:fusionVectorEstimation} presents an overview of our method -- a multimodal representation and estimator based on rank-fusion graphs. The solution is composed of three main generic components, introduced here and detailed in the next sections. Two phases are also defined.
}

\changed{Based on a train set and multiple rankers, a {\em training} phase generates fusion graphs and then fusion vectors for the train set. Then, a multimodal prediction model is built based on the fusion vectors and their corresponding train labels.
The second phase, named {\em inference} phase, refers to the multimodal prediction. It first applies for the test sample a similar rank-based fusion approach for multimodal representation, then predicts an outcome for that test sample.
The training phase is performed only once, while the inference phase is performed per prediction.
The first two components -- {\em fusion graph extraction} and {\em graph embedding} -- are used in both phases.}

\changed{The {\em fusion graph extraction} (component 1 in Fig.~\ref{fig:fusionVectorEstimation}) generates a fusion graph $\FG$ for a given test sample $q$. $\FG$ is an aggregated representation of multiple ranks for $q$ that intends to encode information of multiple ranks.
By doing so, this representation is capable of correlating any object to the others from train set, in an unsupervised manner, with respect to multiple modalities.}
\changed{This formulation is presented in Section~\ref{sec:fusionGraphExtraction}.
{\em Graph Embedding}~(2) projects fusion graphs into a vector space model, producing a corresponding fusion vector $\FV$ for $\FG$, then enabling better performance and efficiency to the representation model.
We propose an embedding formulation in Section~\ref{sec:fusionGraphEmbedding}.
At the end, a multimodal learning component (3) builds an {\em estimator} trained over the response fusion vectors, in order to predict for test samples (also modeled as fusion vectors).
This is detailed in Section~\ref{sec:fusionVectorSearch}.}

\subsection{Rank-Fusion Graphs}
\label{sec:fusionGraphExtraction}
The fusion graph extraction component produces $\FG$ for a given $q$, in terms of $m$ rankers and $n$ response items.
A fusion $\FG$ graph is an encoding of ranks for $q$, aiming at encapsulating and correlating $\rankof{q}$.
\changed{$\FG$ works as an underlying representation model that enables -- by means of multiples modalities and rankers -- multimodal objects to be compared and analyzed in a uniform and robust way.
One $\FG$ is generated for each query sample.}

\changed{We follow the fusion graph formulation from~\cite{Dourado:2019:FG}, referred to as \textit{FG}, that defines a mapping function $q \mapsto \FG$, based on its ranks $\rank \in \ranksof{q}$ and ranks' inter-relationships.
The initial formulation includes a dissimilarity function for $\FG$, and a retrieval model based on fusion graphs. Here, however, we focus on the definition of \textit{FG} itself, extending its use as part of a rank-based late-fusion approach for representation model in multimodal prediction tasks, without these components.}

\begin{figure}
\centering\includegraphics[width=0.3\linewidth]{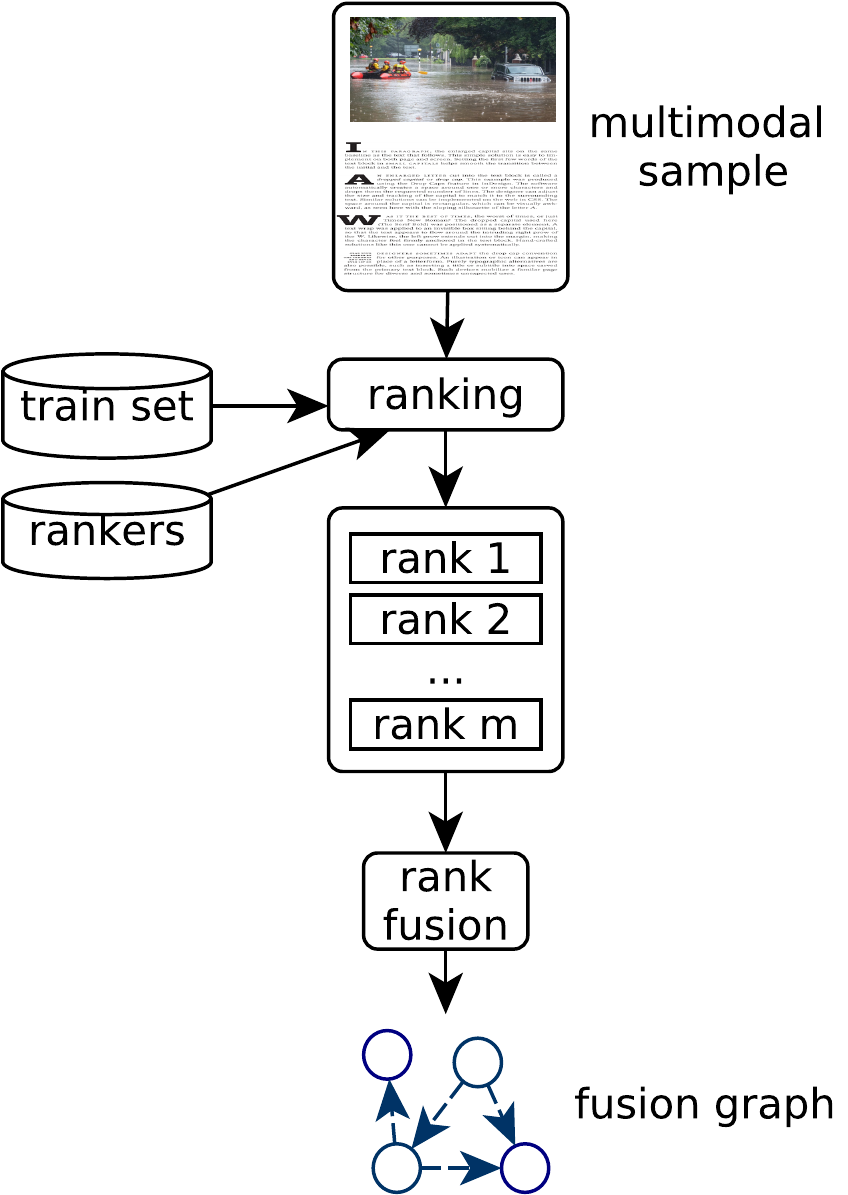}
\caption{Extraction of a fusion graph.}\label{fig:graphFusionFromSample}
\end{figure}

The process is illustrated in Fig.~\ref{fig:graphFusionFromSample}. Given a query (or test sample) $q$, $m$ rankers, and a train set of size $n$, $m$ ranks are generated. These ranks are then normalized to allow for producing the fusion graph $\FG$ for $q$.
In rank normalization, the scores in the ranks generated by dissimilarity-based comparators are converted to similarity-based scores. Besides, all ranks have their scores rescaled to the same interval.

\changed{In this formulation, $\FG$ is built as a weighted directed graph.
$\FG$, for $q$, and it includes all response items from each $\rankof{q} \in \ranksof{q}$ as vertices.
In essence, the vertex set for $\FG$ is composed of the union of all response samples found in all ranks defined for $q$.
A vertex $v_A$ is associated with a response sample $A$.
The vertex weight of $v_A$, expressed by $w_{v_A}$ and given by Equation~\ref{eq:Wv}, is expected to encode how relevant $A$ is to $q$.
$w_{v_A}$ is taken as the sum of the similarities that $A$ has in the ranks of $q$.}

\changed{Edges are created between vertices by taking into account the degree
of relationship of their response items, and the degree of their relationships to $q$.
There is an edge $e_{A,B}$, linking $v_A$ to $v_B$, if $A$ and $B$ are both responses in any rank of $q$ and if $B$ occurs in any rank of $A$.
The weight of $e_{A,B}$, $w_{e_{A,B}}$, is the sum of the similarities that the response item $B$ has in the ranks of $A$, divided by the position of $A$ in each rank of $q$ (Equation~\ref{eq:We}), considering position values starting by $1$.}

\begin{equation}
\label{eq:Wv}
w_{v_A} = \sum_{A \in \rankof{i} \wedge \rankof{i} \in \ranksof{q}} \scorein{\rankof{i}}{q}{A}
\end{equation}	

\begin{equation}
\label{eq:We}
w_{e_{A,B}} = \sum_{ A \in \rankof{i} \wedge \rankof{i} \in \ranksof{q}} \sum_{B \in \rankof{j} \wedge \rankof{j} \in \ranksof{A}} \left(\frac{\scorein{\rankof{j}}{A}{B}}{\rankposition{\rankof{i}}{A}}\right)
\end{equation}

\subsection{Embedding of Rank-Fusion Graphs}
\label{sec:fusionGraphEmbedding}

Let $\FGset=\{\FG_i\}_{i=1}^n$ be the fusion graph set for the response set of a given collection.
Based on $\FGset$, an embedding function $\FGProjector$ defines a vector space in which a fusion graph $\FG$ is projected as a fusion vector $\FV$, i.e., $\FV=\FGProjector(\FG)$ for any $\FG$.
\changed{We claim $\FV$ can be adopted as a suitable representation model of multi-ranked objects for multimodal prediction tasks. It encodes the use of multiple rankers and allows the fusion of multiple modalities.}

\changed{Important advantages arrive from this embedding process to our prediction framework:
(i) the vector domain brings broader availability of algorithms and techniques, compared to the graph domain;
(ii) the objects can be indexed by structures such as search trees or inverted files~\cite{manning:2008:introIR}, promoting overall efficiency;
and (iii) improved compression and storage capabilities can be applied.}

Dissimilarity scores between fusion vectors can be obtained by traditional vector comparators, ranging from correlation metrics to traditional distances and dissimilarity functions, such as Jaccard, cosine, or the Euclidean distance.
Note also that fusion graph vector representations can be combined with indexing mechanisms to allow fast retrieval, in sub-linear time.
$\FGProjector$ can be defined by either supervised or unsupervised or strategies.
We explore three approaches in this paper, preliminarily presented in~\cite{Dourado:2019:FV} in the context of retrieval tasks. \changed{Different from that work, here we extend its use on prediction tasks involving supervised learning.
As our method is flexible in terms of its components, other embeddings could be further explored, based on node signatures~\cite{wong:2006:graphSignature,hu:2014:signatureGraph} or subgraph matching~\cite{jouili:2009:graphMatching,zou:2014:graphQueryEngine}.}

This first embedding approach, $\FGProjector_V$, derives the vector space based on vertex analysis.
Let $w_{\FG}(v)$ be the weight of the vertex $v$, if $v \in \FG$, otherwise 0.
Similarly, let $w_{\FG}(e)$ be the weight of the edge $e$, if $e \in \FG$, otherwise 0. Also, let $d$ be the dimensionality of the vector space induced from $\FGProjector$, such that $\FV \in \real^d$.
\changed{For $\FGProjector_V$, one vector attribute is induced for each response object, therefore $d=|\FV|=|\FGset|$.
$\FGProjector_V$ generates the weights in $\FV$ according to the vertices from $\FG$, such that $\FV[i]$ is the importance value of the $i$-th attribute, therefore $\FV[i]=w_{\FG}(v_i)$.}
Despite the vector space increases linearly to the collection size, the fusion vectors are sparse, i.e., a few non-zero entries compose the vector, which makes this approach simple and efficient in practice.

$\FGProjector_H$ is a hybrid embedding approach based on both vertex and edge analysis. $\FGProjector_H$ encodes more information into the vector space, at a cost of a higher dimensionality $d=\frac{n^2}{2}$.
\changed{These two approaches explore node-based and edge-based local signatures~\cite{wong:2006:graphSignature}. Comparing both, $\FGProjector_{H}$ is expected to gain in effectiveness and lose in efficiency.
$\FV$ is defined as
\begin{equation} \label{eq:fvHybrid}
\FV=(u_1,\ldots,u_i,\ldots,u_n,x_1,\ldots,x_k,\ldots,x_m),
\end{equation}
where $1 \le i \le n$, $1 \le k \le m$, $m=\frac{n^2}{2}-n$,
$u_i=w_{\FG}(v_i)$, $i < j$, and $x_k=w_{\FG}(e_{v_i,v_j})+w_{\FG}(e_{v_j,v_i})$.}

The third approach, $\FGProjector_K$, extends the \ac{BoG} archetype to embed graphs as a histogram of kernels, where the vectorial attributes are selected by unsupervised selection of common subgraph patterns. The kernels are obtained from the centroids of a graph clustering process. Then, a vector quantization process, consisting of assignment and pooling procedures, is adopted to embed an input graph to the vector space.
\changed{$\FGProjector_K$ extends BoG using the following definitions:
\begin{itemize}
\item \ac{GoI}: a kernel sub-structure from $\FG$, such that a set of valid subgraphs from $\FG$ can be induced. For every vertex $v \in \FG$, one undirected connected graph is induced containing: (1) $v$; (2) all direct incident vertices leaving from $v$; and (3) the edges between them. We preserve vertex weights and edge weights into \ac{GoI}.
\item GoI Dissimilarity Function: provides a dissimilarity score between two subgraphs. We employ MCS (Equation~\ref{eq:MCS}), which compares two graph in terms of their maximum common subgraphs and produces a dissimilarity score. Under the assumption of non repetition of vertices within a graph, MCS is linear on the number of vertices~\cite{Dourado:2019:BoTG}.
\item Codebook Generation: a subset of the training GoI's must be selected to compose the codebook. BoG suggests clustering or random selection. We perform a graph clustering using MeanShift~\cite{comaniciu:2002:meanshift}, adapted to work with a distance matrix as input, which we compute with the GoI Dissimilarity Function.
\item Assignment: defines an activation value correlating a subgraph to a vector attribute. We adopt Soft Assignment, which employs a kernel function that establishes, for an input subgraph, a score for each graph attribute~\cite{Silva:2018:BoG}. Soft Assignment is given by Equation~\ref{eq:soft_assignment}~\cite{Dourado:2019:BoTG}, where $S$ is the set of subgraphs for a certain input graph, $a_{ij}$ is the assignment value between the subgraph $s_i \in S$ and the attribute $w_j$, $d$ is the vocabulary size, $D(s_{i},w_{k})$ computes the GoI dissimilarity between $s_i$ and $w_k$, and ${K(x) = \frac{exp(-\frac{x^2}{2\sigma^2})}{\sigma\sqrt{2\pi}}}$ is a Gaussian applied to smooth the dissimilarities~\cite{van:2010:visual}. $\sigma$ allows smoothness control.
\item Pooling: summarizes assignments, producing the final vector. We adopt Average Pooling, which assigns to the $j$-th vector attribute a weights based on the percentage of associations between the input sample subgraphs to the $j$-th graph attribute. This is given by Equation~\ref{eq:avg_pooling}~\cite{Dourado:2019:BoTG}, where $v[j]$ refers to the $j$-th vector attribute.
\end{itemize}}

\begin{equation}
\label{eq:MCS}
dist_{MCS(G_a,G_b)} = 1 - \frac{|mcs(G_a,G_b)|}{max(|G_a|,|G_b|)}
\end{equation}

\begin{equation}
\label{eq:soft_assignment}
a_{ij} = \frac{K(D(s_i, w_j))}{sum_{k=1}^{d} K(D(s_i, w_k))}
\end{equation}

\begin{equation}
\label{eq:avg_pooling}
v_j =  \frac{ \sum_{i=1}^{|S|} a_{ij} }{ |S| }
\end{equation}

\changed{$\FGProjector_K$ is potentially more discriminative and concise than $\FGProjector_V$ and $\FGProjector_{H}$ in terms of the resulting embedding, but it requires more computation, domain specialization, and hyperparameter tuning.}
\ac{BoG} has been successfully applied in scenarios involving graph classification, textual representation, and information retrieval~\cite{Silva:2018:BoG,Dourado:2019:BoTG,Dourado:2019:FV}.
This is so far the first extension of \ac{BoG} in the context of multimodal representation for prediction tasks.

\changed{Among those embeddings, $\FGProjector_{K}$ is expected to be the best in terms of effectiveness, but to be the worst with regard to efficiency. This tradeoff must be evaluated in the target scenario to guide the choice of the type of embedding.}
We refer to FV-V as the fusion vector generated by $\FGProjector_V$, while FV-H is generated by $\FGProjector_H$, and FV-K by $\FGProjector_K$.

\subsection{Multimodal Prediction based on Graph Representations}
\label{sec:fusionVectorSearch}

Let $S$ be a training corpus of size $n$.
\changed{We refer to predictor as a statistical model that learns from historical data to make predictions about future or unknown events~\cite{Mitchell:1997:ML}.}
A predictor can be modeled as $f (X, \beta) \approx Y$, where $f$ is an approximation function, $X$ are the independent variables, $Y$ is the dependent variable (target), and $\beta$ are unknown parameters. A learning model explores $S$ to find an $f$ that minimizes a certain error metric.
The training samples are generally labeled, so $Y$ may be categorical. Still, a regressor can be built if the ground truth is numerical, as $E(Y|X)=f(X, \beta)$, so that posterior probabilities are inferred in order to estimate a confidence score of a sample to refer to a class of not.

\changed{Fusion vectors allow the creation of predictors, such as classifiers or regressors, and also ad-hoc retrieval systems, depending on the underlying demanded task.
In this work, we apply them to build multimodal predictors, where training objects -- associated with ground-truth information -- are used to train a supervised estimator. This estimator infers, for a given input object, whether it relates to a label (or event) or not.
$X$, in our case, refers to the fusion vector, acting as a set of variables that intrinsically describes the sample in terms of its multiple multimodal information and its relationship to $S$.}

\changed{From $S$ and $m$ rankers, we build $\FGset=\{\FG_i\}_{i=1}^n$, from which we derive $\FVset=\{\FV_i\}_{i=1}^n$. A classifier $f$ is trained over $\FVset$ and their corresponding labels. Any standard learning model can be applied, such as multiclass SVM, a neural network, random forests, and so forth. We suggest the adoption of non-linear classifiers that can handle large dimensionality, such as SVM or gradient boosting.}

\subsection{\changed{Cost Analysis}}
\label{costAnalysis}

Base rankers usually adopt indexing structures, such as search trees or inverted files, leading to rank generations in sub-linear time with respect to the response set, taking $O(\log{}n)$. One rank is stored in $O(L)$, due to our hyper-parameter $L$.

The fusion graph extraction (Section~\ref{sec:fusionGraphExtraction}) has an asymptotic cost of $O(mL)$ to induce the vertices, and $O(mLmL)$ to induce the edges, leading to $O(m^2L^2)$. As both small values of $L$ and small number ($m$) of rankers are used, the cost of the graph extraction itself is negligible when compared to the prior generation of base ranks.
The number of vertices in each fusion graph is $O(mL)$ in the worst case. Therefore, the storage of a fusion graph is $O(m^2L^2)$ in the worst case.

As the adopted fusion graph does not repeat vertex labels, we can adopt efficient graph comparison functions based on minimum common subgraphs computed in $O(|V_1||V_2|)$~\cite{dickinson2004matching}. The comparison between two fusion graphs is bounded to $O(m^2L^2)$. Both aspects lead to fairly efficient graph comparators in practice.

The inference phase relies on the existence of ranks, fusion graphs and fusion vectors from the response set plus the estimator.
The rank generation for the response set, considering $m$ rankers, takes $O(n(m\log{n}))$, and requires $O(n(mL))$ for storage.
As for the response fusion graphs, it takes $O(nm^2L^2)$ for both generation and storage.
The embedding of rank-fusion graphs take $O(mLn)$ for the train set and $O(mL)$ for a test sample, as for $\FGProjector_V$, and it depends on the underlying embedding approach.
The cost storage of fusion vectors also depends on the embedding approach, as the dimensionality varies. The vectors produced by the proposed approaches are sparse, and it takes $O(mLn)$ for the train set.
Besides, as they are used for multimodal learning, they do not need to be stored. Only the estimator must be kept.
The estimator can be trained in $O(n)$ by learning models such as SVM. A multimodal prediction for an input fusion vector takes $O(1)$.

The final cost of the training stage is $O(n(m\log{}n + m^2L^2))$ for execution, and $O(nm^2L^2)$ for storage, or approximately $O(n\log{n})$ and $O(n)$ for small values of $m$ and $L$, which is overall efficient.

The cost of an inference is the sum of the costs for: generating the ranks $\ranksof{q}$ for $q$; generating the query fusion graph $G_{\ranksof{q}}$; generating the fusion vector $\FV$; and the prediction by the estimator over $\FV$.
The cost from steps two to four are negligible when compared to the first.
The generation of $\ranksof{q}$ is asymptotically limited by the slowest ranker. In general, this step takes $O(m\log{}n)$.
The second and third steps take $O(m^2L^2)$, and the final step is $O(1)$.
The method is efficient and flexible in terms of its components.
The main trade-off is the eventual reduction in the prediction time in relation to the potential gain in accuracy.

\section{Experimental Evaluation}
\label{secExpEval}

This section presents the experimental protocol used to evaluate the method, and the results achieved comparatively to state-of-the-art baselines.
We evaluate the effectiveness of our method as a representation model in prediction tasks.
The focus is on validating our fusion method comparatively to the individual use of descriptors with no fusion, as well as to compare it to early-fusion and late-fusion approaches.

\subsection{Evaluation Scenarios}

We evaluate the proposed method on multiple datasets, comprised of heterogeneous multimodal data, in order to assess its general applicability.
Our experimental evaluation comprises four scenarios, as follows:

\begin{itemize}
\item \textbf{ME17-DIRSM} dataset, the acronym for MediaEval 2017 Disaster Image Retrieval from Social Media~\cite{bischke:2017:meSatellite}, is a multimodal dataset of a competition whose goal is to infer whether images and/or texts refer to flood events or not.
The samples contain images along with textual metadata, such as title, description, and tags, and they are labeled as  either flood (1) or non-flood (0). The task predefines a development set (devset) of \num{5280} samples, and a test set of \num{1320} samples, as well as its own evaluation protocol.

\item \textbf{Brodatz}~\cite{brodatz:1966} is a dataset of texture images, labeled across $111$ classes. There are $16$ samples per class, composing a total of \num{1776} samples.

\item \textbf{Soccer}~\cite{van:2006:soccer} is an image dataset, labeled across 7 categories (soccer teams), containing $40$ images each.

\item \textbf{UW}~\cite{Deselaers:2008:UW}, or University of Washington dataset, is a multimodal collection of \num{1109} images annotated by textual keywords. The images are pictures labeled across $22$ classes (locations).
Pictures per class
vary from 22 to 255. The number of keywords per picture vary from 1 to 22.
\end{itemize}

\subsection{Evaluation Protocol}

For any dataset that does not explicitly define train and test sets, we initially split it into stratified train and test sets, at a proportion of 80\%/20\%. The proportions per class remain equal. The same train and test sets per dataset are adopted to evaluate all methods under the same circumstances, as well as the evaluation metrics.

For each representation model, we fit a multiclass SVM classifier, with the one-vs-all approach and linear kernel, as it is a good fit for general applicability.
Hyper-parameters are selected by grid search on the train set, using 5-fold cross validation.
\changed{We evaluate the effectiveness of each method by the balanced accuracy score -- defined as the average of recall scores per class -- which is suitable to evaluate on both balanced and imbalanced datasets.
The methods are compared by their balanced accuracy.}

The descriptors compose rankers, which are employed to generate ranks in our late-fusion representation. Our method varies with respect to which rankers are used, whether visual or textual rankers, or even their combinations for the multimodal scenario are applied. Besides that, it also varies with respect to which embedding approach is adopted. We evaluate these aspects experimentally.

We model our solution as a rank-fusion approach, followed by an estimator based on rank-fusion vectors. This approach intends to validate our hypothesis that unsupervised graph-based rank-fusion approaches can lead to effective representation models for prediction tasks in general.

We adopt the same experimental evaluation for all datasets but ME17-DIRSM, that defines its own procedure. In this case, the task imposes three evaluation scenarios, as follows.
In the first one, called ``visual'', only visual data can be used.
In the second, called ``textual'', only textual data are used.
In the third, called ``multimodal'', both visual and textual data are expected to be used.
The correctness is evaluated, over the test set, by the metric Average Precision at $K$ (AP@K) at various cutoffs (e.g., 50, 100, 250, and 480), and by their mean value (mAP).

Although the ME17-DIRSM task may be seen as a multimodal binary classification problem, the evaluation metrics require ranking-based solutions, or equivalently confidence-level regressors, so that the first positions are the most likely to refer to a flood event.
For the estimator component in ME17-DIRSM, we adopt SVR, an L2-regularized logistic regression based on linear SVM in its dual form, with probabilistic output scores, and trained over the fusion vectors from devset.
Probabilistic scores are used so that we can sort the test samples by confidence expectancy of being flood.

In ME17-DIRSM, our results are compared to those from state-of-the-art baselines.
In Soccer, Brodatz, and UW, our results are compared to those from two major fusion approaches: concatenation, and majority vote. They cover baselines from both early-fusion and late-fusion families.
For the concatenation procedure, we first normalize the vectors to the $[0,1]$ interval for each attribute, in order to avoid disparities due to different descriptor attribute ranges.
We apply majority voting in the scenarios involving an odd number of descriptors, so that each predicted class is taken as the one most frequently predicted by the estimators constructed for each descriptor.

\subsection{Descriptors and Rankers}

\begin{table*}
\caption{Datasets and descriptors for the experimental evaluation.}
\centering\label{tab:datasets}
\begin{tabular}{llX{10cm}}\hline
Dataset & Data & Descriptors \\\hline
ME17-DIRSM & pictures, textual metadata & ResNet50IN, VGG16P365, NASNetIN, BoW, 2grams, doc2vecWiki \\
Brodatz & texture images & \ac{JCD}, \ac{FCTH}, \ac{CCOM} \\
Soccer & pictures & \ac{BIC}, \ac{GCH}, \ac{ACC} \\
UW & pictures, textual keywords & \ac{JAC}, \ac{ACC}, \ac{JCD}, \ac{CEDD}, word2vecSum, word2vecAvg, doc2vecWiki, doc2vecApnews \\
\hline
\end{tabular}
\end{table*}


For ME17-DIRSM, we selected three visual descriptors and three textual descriptors, for individual analysis in the designed evaluation scenarios, and to evaluate different possibilities of rank-fusion aggregations. We adopt the following state-of-the-art visual descriptors: (i) \textit{ResNet50IN}: 2048-dimensional average pooling of the last convolutional layer of ResNet50~\cite{he:2016:ResNet50}, pre-trained on ImageNet~\cite{Russakovsky:2015:ImageNet}, a dataset of about 14M images labeled for object recognition; (ii) \textit{VGG16P365}: 512-dimensional average pooling of the last convolutional layer of VGG16~\cite{simonyan:2014:VGG16}, pre-trained on Places365-Standard~\cite{zhou:2017:Places}, a dataset of about 10M images of labeled scenes; (iii) \textit{NASNetIN}: 2048-dimensional average pooling of the last convolutional layer of NASNet~\cite{barret:2017:NASNet}, pre-trained on ImageNet dataset.

Based on the textual metadata in ME17-DIRSM, we adopt the following descriptors: (i) \textit{BoW}: Bag of Words (BoW) with Term Frequency (TF) weighting; (ii) \textit{2grams}: 2grams with TF weighting; (iii) \textit{doc2vecWiki}: 300-dimensional doc2vec~\cite{le:2014:doc2vec} pre-trained on English Wikipedia dataset, of about 35M documents and dumped at 2015-12-01.

For the other datasets, we elected a number of heterogeneous descriptors. In Soccer: \ac{BIC}, \ac{GCH}, and \ac{ACC}. In Brodatz: \ac{JCD}, \ac{FCTH}, and \ac{CCOM}. In UW: \ac{JAC}, \ac{ACC}, \ac{JCD}, and \ac{CEDD}, as visual descriptors, and \textit{word2vecSum}, \textit{word2vecAvg}, \textit{doc2vecWiki}, and \textit{doc2vecApnews}, as textual descriptors.

For the deep networks used for CNN-based visual feature extraction, as well as in the textual feature extraction with doc2vec, we take advantage of pre-trained models. This practice, known as transfer learning, has been effective in many scenarios~\cite{kornblith:2018:imagenetModelTransfer}, and it is also particularly beneficial for datasets that are not large enough to generalize the training of such large architectures, as in our case. Because the problem requires prediction of flood images, we prioritize, in the selection of visual descriptors, datasets for pre-training that focus on images of scenes, aiming at better generality to the target problem.

We perform the same preprocessing steps for every textual descriptor: lower case conversion, digit and punctuation removal, and English stop word removal. For \textit{BoW} and \textit{2grams}, we also apply Porter stemming.

The \textit{word2vecSum} descriptor produces, for any input document, a vector corresponding to the sum of the word embedding vectors~\cite{mikolov:2013:word2vec} related to each term within that document, while \textit{word2vecAvg} computes the mean vector of them.

The doc2vec model promotes document-level embeddings for texts, and it is based on word embeddings~\cite{mikolov:2013:word2vec}, a preliminary work that assigns vector representations for words in order to capture their semantic relationships. \textit{doc2vecApnews} stands for a 300-dimensional \textit{doc2vec}~\cite{le:2014:doc2vec} model, pre-trained over the Associated Press News textual dataset, of about 25M news articles from 2009 to 2015.
The datasets and descriptors are summarized in Table~\ref{tab:datasets}.

From the descriptors, rankers are defined as tuples of (descriptor, comparator), where the comparator corresponds to a dissimilarity function.
We compose a ranker for each descriptor by choosing an appropriate comparator.
We define dissimilarity functions to be used along with those descriptors that are not explicitly associated with one.
This is the case for the textual descriptors adopted in UW, as well as the descriptors adopted in ME17-DIRSM. All remaining descriptors define their own comparators.
For the textual descriptors BoW and 2grams, we adopt the weighted Jaccard distance, defined as $1 - J(\mathbf{u}, \mathbf{v})$, where $J$ is the Ruzicka similarity metric (Equation~\ref{eq:Jaccard}).
Jaccard is a well-known and widely-used comparison metric for classic textual descriptors, especially for short texts, as in our case.
For the remaining descriptors, we choose the Pearson correlation distance, defined as $1 - \rho(\mathbf{u}, \mathbf{v})$ (Equation~\ref{eq:correlation}), which is a general-purpose metric due to its suitability for high dimensional data and scale invariance.
\begin{equation}\label{eq:Jaccard}
J(\mathbf{u},\mathbf{v}) = \frac{\sum_i \min(u_i,v_i)}{\sum_i \max(u_i,v_i)}
\end{equation}
\begin{equation}\label{eq:correlation}
\rho(\mathbf{u}, \mathbf{v}) = \frac{(\mathbf{u} - \bar{u}) \cdot (\mathbf{v} - \bar{v})}
{{\|(\mathbf{u} - \bar{u})\|}_2 {\|(\mathbf{v} - \bar{v})\|}_2}
\end{equation}


\subsection{Fusion Setups}

For both visual and textual scenarios in ME17-DIRSM, we analyze three variants of our method with respect to the input rankers for late-fusion.
For the visual scenario, the combinations are
ResNet50IN + NASNetIN,
ResNet50IN + VGG16P365,
and ResNet50IN + NASNetIN + VGG16P365.
For the textual scenario, the combinations are
BoW + 2grams,
BoW + doc2vecWiki,
and BoW + 2grams + doc2vecWiki.

As for the multimodal scenario, we investigate some combinations taking one ranker of each type, two of each, and three of each.
Six multimodal combinations are evaluated:
ResNet50IN + BoW,
ResNet50IN + NASNetIN + BoW + 2grams,
ResNet50IN + NASNetIN + BoW + doc2vecWiki,
ResNet50IN + VGG16P365 + BoW + 2grams,
ResNet50IN + VGG16P365 + BoW + doc2vecWiki,
and ResNet50IN + NASNetIN + VGG16P365 + BoW + 2grams + doc2vecWiki.

We report three results for the adoption of \textit{FV}, in its different embedding approaches, as a representation model for prediction tasks, in Soccer, Brodatz, and UW.
We report the results for multiple descriptor combinations, in order to analyze: (i) the method against baselines, (ii) the embedding approaches, and (iii) the comparative effectiveness between the descriptor combinations. 
The descriptor combinations selected, although not exhaustive, are targeted for a large number of scenarios. In Soccer and Brodatz, all possible combinations were selected for evaluation. In UW, several visual combinations and multimodal combinations were selected.
The descriptor combinations in Soccer are: ACC + BIC, BIC + GCH, ACC + GCH, and ACC + BIC + GCH. In Brodatz: CCOM + FCTH, CCOM + JCD, FCTH + JCD, and CCOM + FCTH + JCD.
In UW: ACC + CEDD, ACC + JCD, CEDD + JAC, CEDD + JCD, ACC + CEDD + JCD, and CEDD + JAC + JCD, for visual fusion, and ACC + doc2vecApnews, JCD + doc2vecApnews, ACC + JCD + doc2vecApnews, ACC + JCD + doc2vecWiki, ACC + JCD + word2vecAvg, and ACC + JCD+word2vecSum, for multimodal fusion.

\subsection{Results and Discussion}

\subsubsection{Base Results}

\begin{figure}
\centering\includegraphics[width=0.6\linewidth]{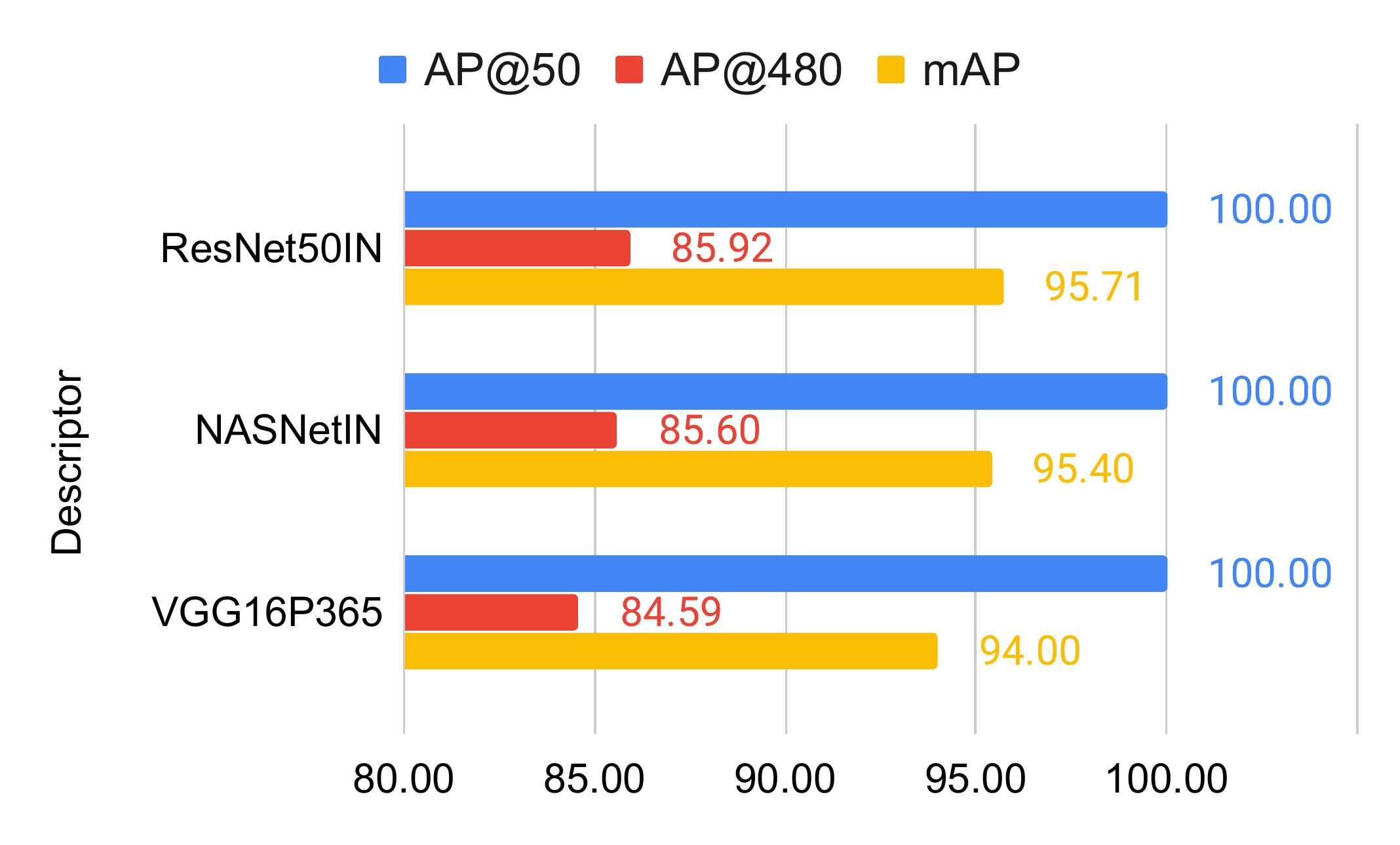}
\caption{Base results of the descriptors, along with a SVR, in ME17-DIRSM, visual scenario.}\label{fig:resultsVisual}
\end{figure}

\begin{figure}
	\centering\includegraphics[width=0.6\linewidth]{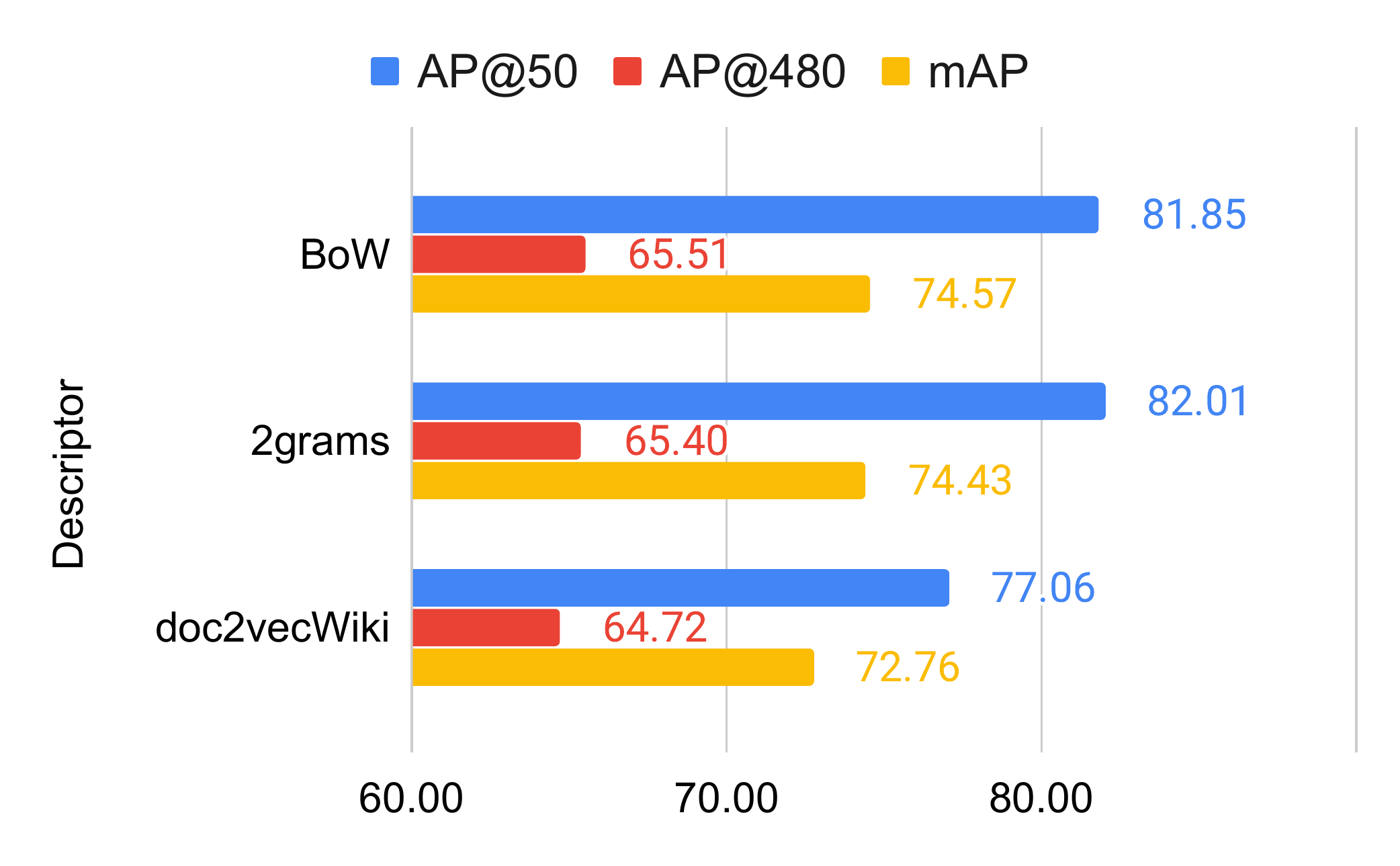}
	\caption{Base results of the descriptors, along with a SVR, in ME17-DIRSM, textual scenario.}\label{fig:resultsTextual}
\end{figure}

Here we report results by the use of individual descriptors. They constitute an initial baseline for our method as well as for other fusion approaches.
Fig.~\ref{fig:resultsVisual} and~\ref{fig:resultsTextual} present the results for the visual and textual scenarios in ME17-DIRSM, achieved by the visual and textual selected descriptors, along with a SVR regressor.
As the task only showed AP@480 and mAP in their leaderboard, we focus our discussions on these two metrics.
The correctness for the visual scenario is already high within these baselines, around $85\%$ in AP@480. In the textual scenario, AP@480 is around $65\%$, which suggests more room for improvement.
We report, in Tables~\ref{tab:descriptorResultsSoccer}, \ref{tab:descriptorResultsBrodatz}, \ref{tab:descriptorResultsUWvisual}, and \ref{tab:descriptorResultsUWtextual}, the results obtained in Soccer, Brodatz, UW (visual), and UW (textual), respectively, for the base descriptors along with a SVM classifier.

\begin{table}
\caption{Base results of the descriptors, along with a SVM, in Soccer.}
\centering\label{tab:descriptorResultsSoccer}
\begin{tabular}{lr}\hline
Descriptor & Balanced acc. \\\hline
BIC & 60.84 \\
GCH & 56.38 \\
ACC & 51.79 \\
\hline
\end{tabular}
\end{table}

\begin{table}
\caption{Base results of the descriptors, along with a SVM, in Brodatz.}
\centering\label{tab:descriptorResultsBrodatz}
\begin{tabular}{lr}\hline
Descriptor & Balanced acc. \\\hline
JCD	& 78.65 \\
FCTH & 75.10 \\
CCOM & 74.65 \\
\hline
\end{tabular}
\end{table}

\begin{table}
\caption{Base results of the descriptors, along with a SVM, in UW, visual scenario.}
\centering\label{tab:descriptorResultsUWvisual}
\begin{tabular}{lr}\hline
Descriptor & Balanced acc. \\\hline
JAC  & 83.74 \\
ACC  & 77.06 \\
JCD  & 73.99 \\
CEDD & 72.46 \\\hline
\end{tabular}
\end{table}

\begin{table}
\caption{Base results of the descriptors, along with a SVM, in UW, textual scenario.}
\centering\label{tab:descriptorResultsUWtextual}
\begin{tabular}{lr}\hline
Descriptor & Balanced acc. \\\hline
word2vecSum & 86.91 \\
word2vecAvg & 82.96 \\
doc2vecEnwi & 82.95 \\
doc2vecApnews & 82.94 \\
\hline
\end{tabular}
\end{table}

\subsubsection{Parameter Analysis}

We begin the experimental evaluation of the method by evaluating its hyper-parameters and proposed variants.
The resulting size of $\FG$ is affected by the input rank sizes, defined by the hyper-parameter \textit{L}. For the same reason, larger \textit{FG}'s either increase the vocabulary sizes of \textit{FV} or the complexity to generate them.
Dourado et al.~\cite{Dourado:2019:FG} showed that an increase in \textit{L} leads to more discriminate graphs up to a saturation point. A practical upper bound for the choice of \textit{L} tends to be the maximum rank size of users' interest, indirectly expressed here by the evaluation metrics.

\begin{figure}
\centering\includegraphics[width=0.8\linewidth]{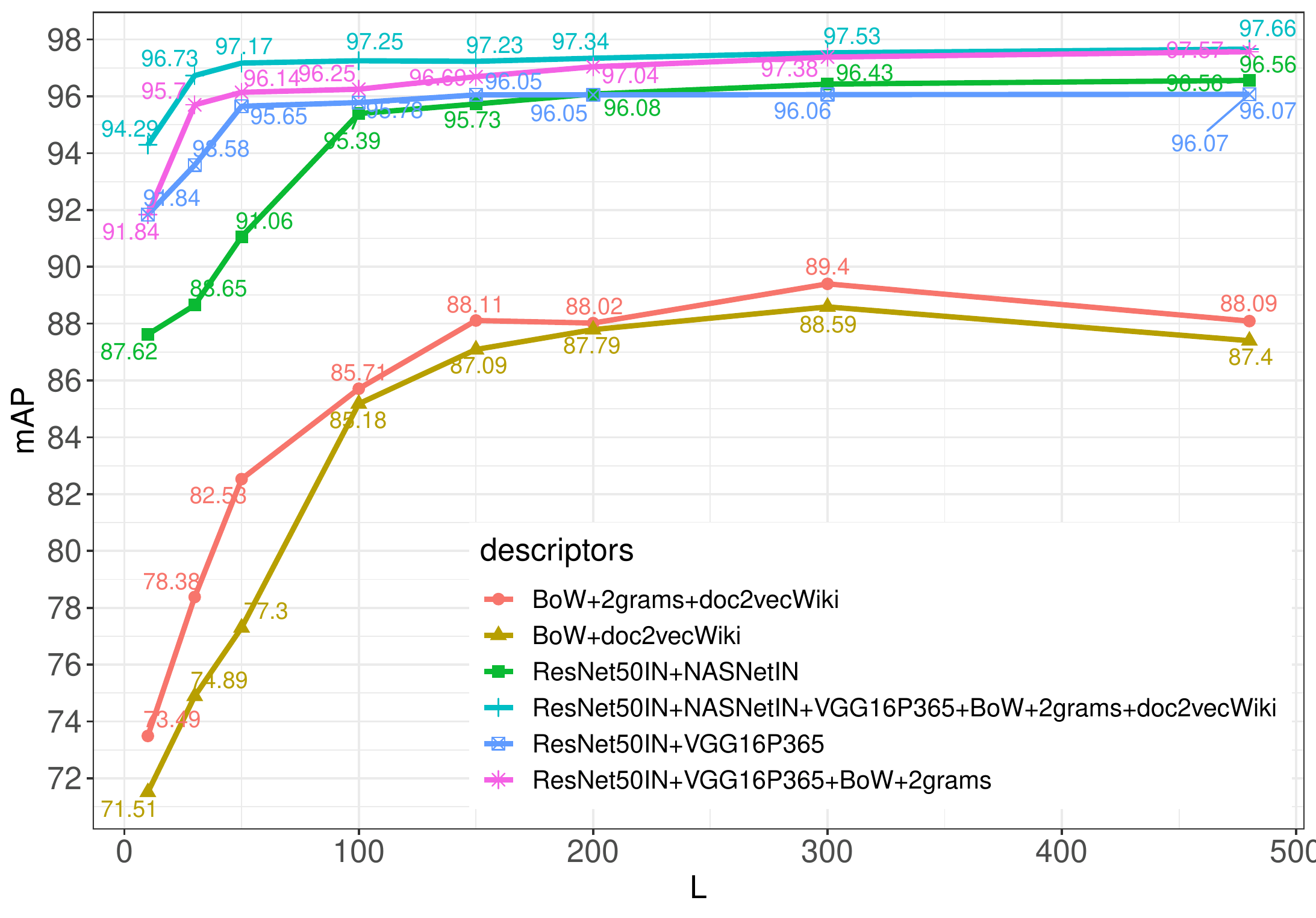}
\caption{Effect of the rank size limit (\textit{L}) for the fusion graph extraction, in the mAP score, for different fusion scenarios in ME17-DIRSM.}\label{fig:parameterEvaluationL}
\end{figure}

As ME17-DIRSM defines evaluation metrics for ranks up to 480, we start by empirically evaluating the influence of \textit{L} in the mAP score, for values to up 480.
Fig.~\ref{fig:parameterEvaluationL} reports the influence of L for some of the elected fusion scenarios. The results were as expected: the effectiveness usually increased as \textit{L} was larger.
For the next evaluation scenarios in ME17-DIRSM, we adopt $L=480$.
For the other datasets, we adopt $L=10$.

\subsubsection{Fusion Results in Event Detection}

\begin{figure}
\centering\includegraphics[width=0.7\linewidth]{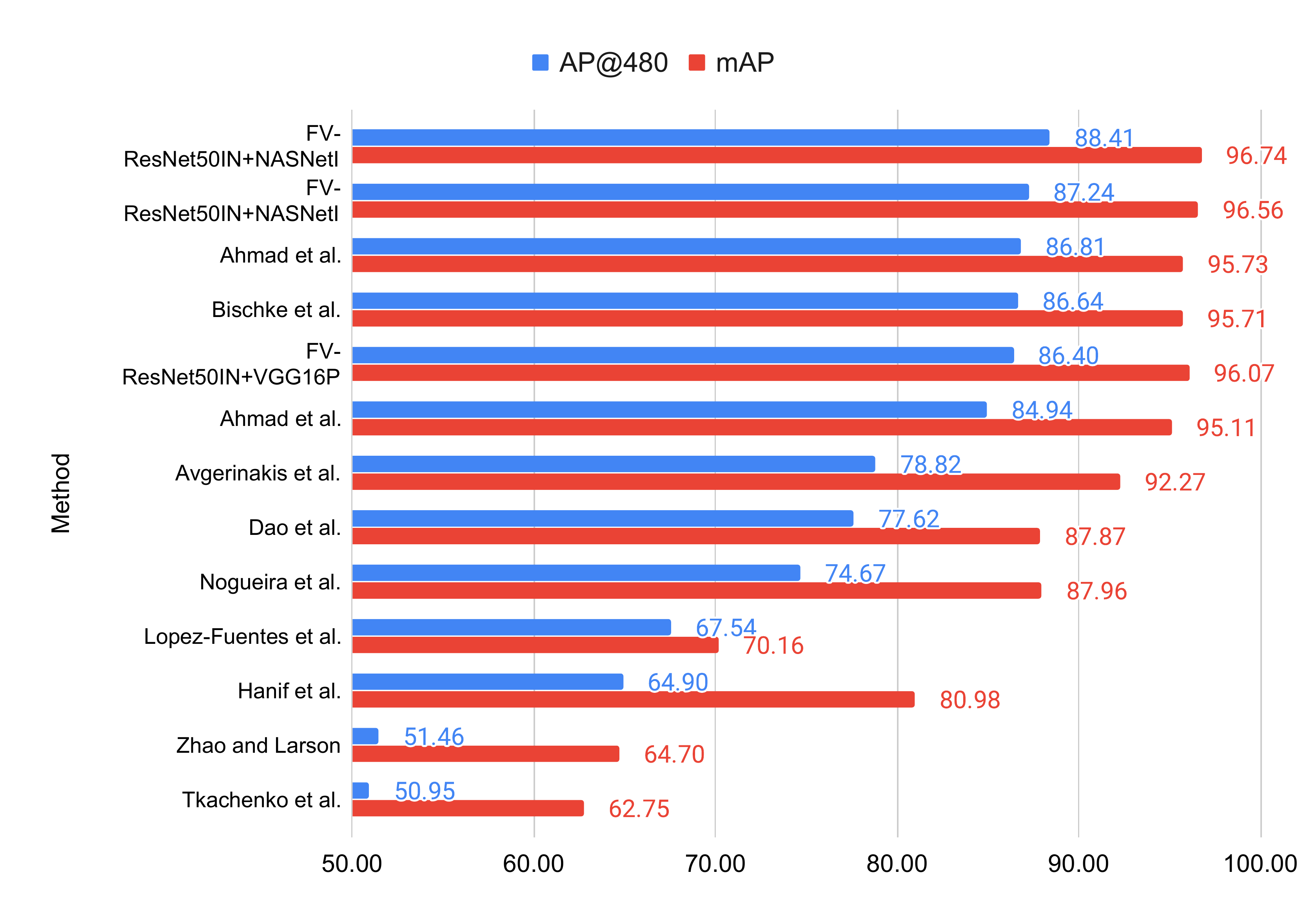}
\caption{Flood detection based on {\bf visual} features, in ME17-DIRSM.}\label{fig:resultsRunVisual}
\end{figure}

\begin{figure}
\centering\includegraphics[width=0.7\linewidth]{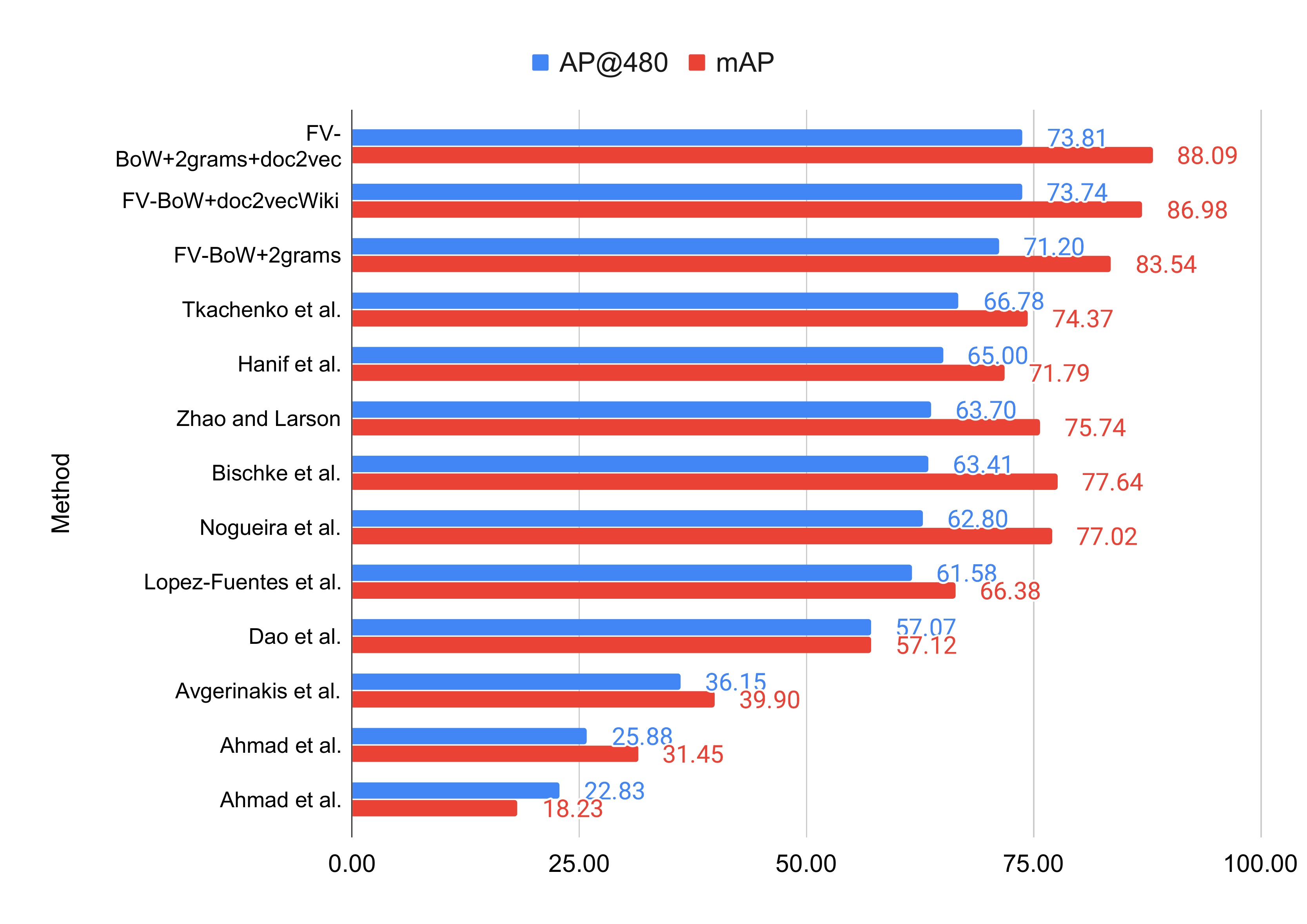}
\caption{Flood detection based on {\bf textual} features, in ME17-DIRSM.}\label{fig:resultsRunTextual}
\end{figure}

\begin{figure}
\centering\includegraphics[width=0.8\linewidth]{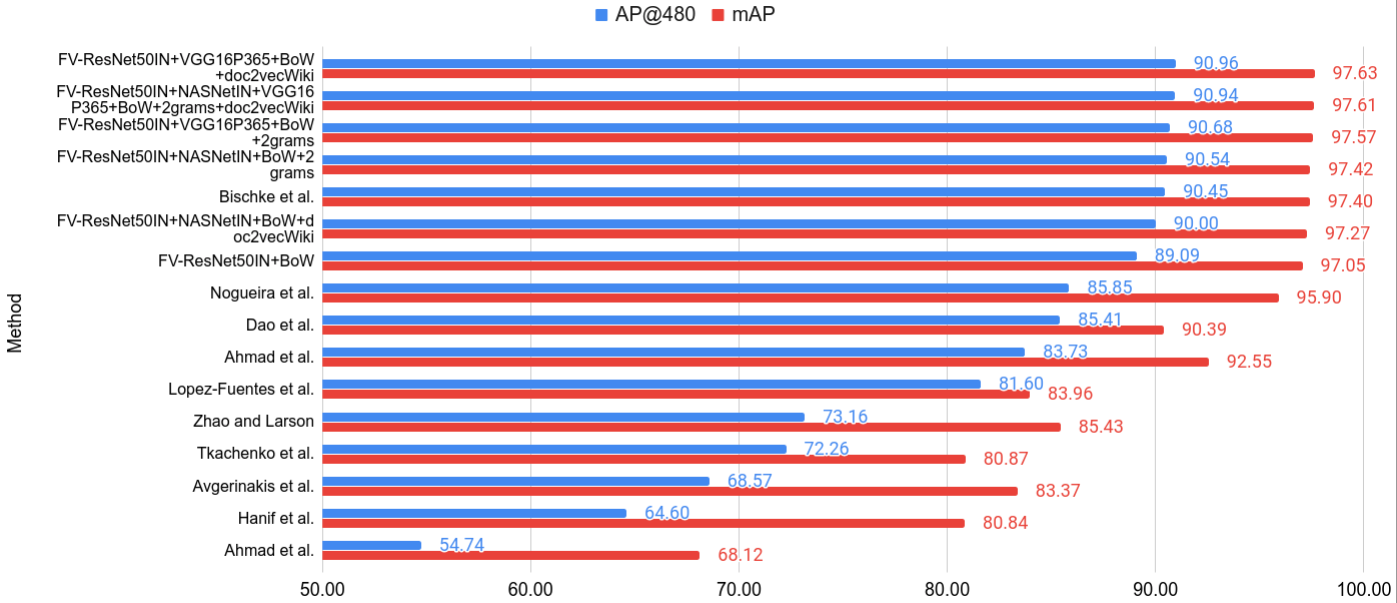}
\caption{Flood detection based on {\bf multimodal} features, in ME17-DIRSM.}\label{fig:resultsRunMultimodal}
\end{figure}

We present our results achieved for the three scenarios in ME17-DIRSM, using the combinations proposed, along with the results of the teams that participated in the competition~\cite{dirsm17:XFuEtAl,dirsm17:SAhmadEtAl,dirsm17:BBischkeEtAl,dirsm17:KAhmadEtAl,dirsm17:KAvgerinakisEtAl,dirsm17:MSDaoEtAl,dirsm17:LLopezFuentesEtAl,dirsm17:MHanifEtAl,dirsm17:ZZhao,dirsm17:NTkachenkoEtAl,dirsm17:KNogueiraEtAl}.
These results are presented in Fig.~\ref{fig:resultsRunVisual}, \ref{fig:resultsRunTextual}, and~\ref{fig:resultsRunMultimodal}, respectively for the visual, textual, and multimodal scenarios.
In ME17-DIRSM, we focused on the $\FGProjector_V$ embedding approach. For the other datasets, we evaluate all of them.

In the visual scenario, only Ahmad et al.~\cite{dirsm17:SAhmadEtAl} and Bischke et al.~\cite{dirsm17:BBischkeEtAl} performed better, in terms of AP@480 and mAP, than our preliminary base setup, based on individual descriptors along with the SVR regressor.
As for the textual scenario, only Tkachenko et al.\cite{dirsm17:NTkachenkoEtAl} surpassed BoW + SVR in AP@480, and only Bischke et al.~\cite{dirsm17:BBischkeEtAl}, Nogueira et al.~\cite{dirsm17:KNogueiraEtAl}, and Zhao and Larson~\cite{dirsm17:ZZhao} surpassed BoW + SVR in mAP.
This indicates that descriptors properly selected to the target problem can overcome more complex models, also requiring less effort.

Our method was superior in the visual scenario to all baselines, for two of three proposed variants of ranker combinations. Compared to the strongest baselines considering this scenario, it presents gains from around 1 to 2\% in AP@480, and 1\% in mAP.
Compared to the visual base results, from the individual descriptors, 3 to 4\% in AP@480, and 1 to 2\% in mAP.

In the textual scenario, our gains were expressive. It was superior in the textual scenario to all related works, for all three proposed variants of ranker combinations. Compared to the strongest baselines, it presents gains from 5 to 7\% in AP@480, and 6 to 11\% in mAP.
Compared to the textual base results, from the individual descriptors, 6 to 8\% in AP@480, and 14 to 16\% in mAP.
In the multimodal scenario, considered baselines were more competitive. Again, however, our method presents gains over them, of 0.5\% in AP@480 and 0.23\% in mAP.


\subsubsection{Fusion Results in Classification Tasks}

Tables~\ref{tab:fusionResultsSoccer}, \ref{tab:fusionResultsBrodatz}, \ref{tab:fusionResultsUWvisual}, and~\ref{tab:fusionResultsUWmultimodal} report the results obtained by our method variants, respectively in Soccer, Brodatz, UW for image data, and UW for multimodal data, besides the results obtained by the baselines.

\begin{table}
\caption{Balanced accuracies by fusion methods in Soccer.}
\centering\label{tab:fusionResultsSoccer}
\begin{tabular}{lrrrr}
\hline
Method        & ACC+BIC   & BIC+GCH   & ACC+GCH   & ACC+BIC+GCH \\\hline
FV-K   & 67.22 & \textbf{67.09} & 66.33 & \textbf{68.88}   \\
FV-V   & \textbf{67.48} & 64.29 & \textbf{67.86} & 68.62   \\
FV-H   & 66.58 & 66.07 & 64.29 & 66.71   \\
concatenation & 62.50 & 62.76 & 55.48 & 63.39   \\
majorityVote  & ---        & ---        & ---        & 62.76   \\\hline
\end{tabular}
\end{table}

\begin{table}
\caption{Balanced accuracies by fusion methods in Brodatz.}
\centering\label{tab:fusionResultsBrodatz}
\begin{tabular}{lrrrr}
\hline
Method             & CCOM+FCTH & CCOM+JCD  & FCTH+JCD  & CCOM+FCTH+JCD \\\hline
FV-H               & \textbf{89.59} & \textbf{91.84} & \textbf{88.13} & \textbf{91.65}     \\
FV-V               & 88.56 & 90.98 & 87.29 & 91.10     \\
FV-K               & 88.37 & 89.15 & 85.69 & 89.07     \\
concatenation & 88.16 & 89.18 & 86.73 & 88.76     \\
majorityVote              & ---        & ---        & ---        & 83.42    \\\hline
\end{tabular}
\end{table}

\begin{table*}
\caption{Balanced accuracies for visual fusion in UW.}
\centering\label{tab:fusionResultsUWvisual}
\begin{tabular}{lrrrrR{2.3cm}R{2.3cm}}
\hline
              & ACC + CEDD     & ACC + JCD      & CEDD + JAC     & CEDD + JCD     & ACC + CEDD + JCD & CEDD + JAC + JCD \\\hline
FV-K          & \textbf{85.88} & \textbf{85.02} & \textbf{84.83} & 73.97          & \textbf{83.97}      & \textbf{85.73}      \\
FV-H          & 84.58          & 84.95 & 84.47  & 74.13          & 80.50          & 81.66      \\
FV-V          & 83.57          & 84.26 & 82.42  & \textbf{74.16}          & 83.71 & 81.87      \\
concatenation & 82.75          & 82.64 & 77.70  & 70.28          & 83.18          & 82.23      \\
majorityVote  & ---            & ---            & ---            & ---       & 76.22      & 73.05     \\\hline
\end{tabular}
\end{table*}

\begin{table*}
\caption{Balanced accuracies for multimodal fusion in UW.}
\centering\label{tab:fusionResultsUWmultimodal}
\resizebox{.99\textwidth}{!}{\begin{tabular}{lR{2.4cm}R{2.4cm}R{2.4cm}R{2.4cm}R{2.4cm}R{2.4cm}}
\hline
 & ACC + doc2vecApnews & JCD + doc2vecApnews & ACC + JCD + doc2vecApnews & ACC + JCD + doc2vecWiki & ACC + JCD + word2vecAvg & ACC + JCD + word2vecSum \\\hline
FV-V               & 86.72           & \textbf{87.21}           & 90.04               & 89.87               & \textbf{92.84}             & \textbf{92.84}             \\
FV-K               & 88.52           & 85.52           & \textbf{90.64}               & \textbf{92.18}               & 92.51             & 92.65             \\
FV-H               & \textbf{90.26}           & 85.44           & 89.56               & 90.86               & 91.73             & 91.73             \\
concatenation & 88.69           & 86.57           & 90.38               & 88.28               & 90.73             & 90.87             \\
majorityVote              & ---             & ---             & 83.77               & 83.11               & 84.62             & 84.49            \\\hline
\end{tabular}}
\end{table*}

Our method led to significant gains when compared to the best base result from the descriptors in each dataset: around 8 p.p. in Soccer, 13.2 p.p. in Brodatz, 2.2 p.p. in UW for visual fusion, and 5.9 p.p. in UW for multimodal fusion. 
The gains in UW were comparatively lower than others, yet consistent, because the base results were already higher, so that there was less room for improvement.
Our method, when compared to the best baseline in each descriptor combination in each dataset, had gains in all cases: up to 12.4 p.p. in Soccer, up to 2.9 p.p. in Brodatz, up to 7.1 p.p. in UW for visual fusion, and up to 3.9 p.p. in UW for multimodal fusion.
Overall, all the FV approaches outperformed all baselines in all datasets.
In only 6 out of 20 descriptor combinations, some of the baselines surpassed any of the FV approaches.

The accuracy disparities, obtained by FV or any other fusion method, across the multiple descriptor combinations in each dataset, show that there is no obvious choice when dealing with which descriptors to be used together. As our representation model is meant to be unsupervised, this choice could only be guided by general heuristics, such as a selection of effective and low correlated descriptors, as discussed in~\cite{Dourado:2019:FG}. We leave this exploration for future work.

FV-K usually outperformed FV-H and FV-V, and FV-H usually outperformed FV-V, although in both cases the gains were at most 3 p.p. On the opposite side, FV-V is the simplest among the three, and FV-K requires more computational steps. These two aspects combined impose that the practical choice among the three must take into account the trade-off between accuracy vs computational cost. In any case, our method is unsupervised and feasible for general applicability.

\section{Conclusions}
\label{secConc}

This paper presented an unsupervised graph-based rank-fusion approach as a representation model for multimodal prediction tasks. Our solution is based on encoding multiple ranks into a graph representation, which is later embedded into a vectorial representation. Next, an estimator is built to predict if an input multimodal object refers to a target event or not, given their graph-based fusion representations.
The proposed method extends the fusion graphs -- first introduced in~\cite{Dourado:2019:FG} -- for supervised learning tasks. It also applies a graph embedding mechanism in order to obtain the fusions vectors, a late-fusion vector representation that encodes multiple ranks and their inter-relationships automatically.
Performed experiments in multiple prediction tasks, such as flood detection and multimodal classification, demonstrate that our solution leads to highly effective results overcoming state-of-the-art solutions from both early-fusion and late-fusion approaches.
Future work focuses on investigating the impact of semi-supervised and supervised approaches for the fusion graph and fusion vector constructions.
We also plan to investigate our method in other multimodal problems, such as recommendation and hierarchical clustering.
Finally, we plan to evaluate the proposal for other multimedia data, such as audio and video.

\bibliography{bib}

\end{document}